\pdfoutput=1

\documentclass[11pt]{article}

\usepackage[final]{acl}
\usepackage{times}
\usepackage{latexsym}
\usepackage{amsmath}
\usepackage[T1]{fontenc}
\usepackage[utf8]{inputenc}
\usepackage{microtype}
\usepackage{inconsolata}
\usepackage{booktabs}
\usepackage{multirow}
\usepackage{graphicx}
\usepackage{amssymb}
\usepackage{wrapfig}
\usepackage{floatflt}
\usepackage{floatrow}
\usepackage[most]{tcolorbox}
\usepackage{url}

\usepackage{colortbl}

\title{\raisebox{-0.3\height}{\includegraphics[height=1cm]{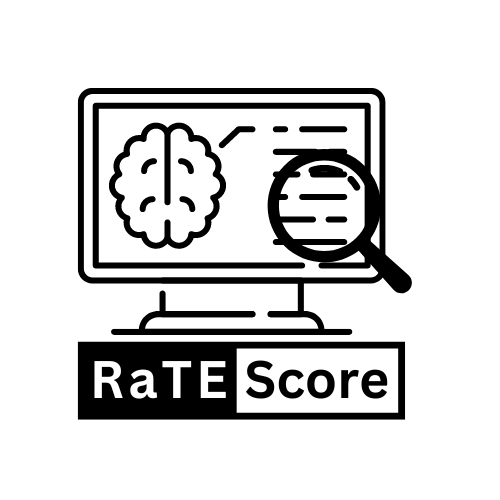}} RaTEScore: A Metric for Radiology Report Generation}

\author{
 \textbf{Weike Zhao\textsuperscript{1,2}},
 \textbf{Chaoyi Wu\textsuperscript{1,2}},
 \textbf{Xiaoman Zhang\textsuperscript{1,2}},
 \\[3pt]
 \textbf{Ya Zhang\textsuperscript{1,2}},
 \textbf{Yanfeng Wang\textsuperscript{1,2,$\dag$}},
 \textbf{Weidi Xie\textsuperscript{1,2,$\dag$}},
\\[3pt]
 \textsuperscript{1}School of Artificial Intelligence, Shanghai Jiao Tong University
\\[3pt]
 \textsuperscript{2}Shanghai Artificial Intelligence Laboratory
\\[3pt]
 \texttt{
 \{zwk0629,wtzxxxwcy02,xm99sjtu,ya\_zhang,wangyanfeng622,weidi\}@sjtu.edu.cn
 }
 \\[3pt]
\href{https://angelakeke.github.io/RaTEScore/}{https://angelakeke.github.io/RaTEScore/}
}

\begin{document}
\maketitle

\begin{abstract}

This paper introduces a novel, entity-aware metric, 
termed as \textbf{Ra}diological Report~(\textbf{T}ext) \textbf{E}valuation~(\textbf{RaTEScore}), to assess the quality of medical reports generated by AI models. RaTEScore emphasizes crucial medical entities, such as diagnostic outcomes and anatomical details. Moreover, it is robust against medical synonyms and sensitive to negation expressions. Technically, we developed a comprehensive medical NER dataset, \textbf{RaTE-NER}, and trained an NER model specifically for this purpose. 
This model enables the decomposition of complex radiological reports into constituent medical entities. The metric itself is derived by comparing the similarity of entity embeddings, obtained from a language model, 
based on their types and relevance to clinical significance. Our evaluations demonstrate that RaTEScore aligns more closely with human preference than existing metrics, validated both on established public benchmarks and our newly proposed \textbf{RaTE-Eval} benchmark.

\end{abstract}

\vspace{-8pt}
\section{Introduction}
\vspace{-2pt}

With the general advancement in natural language processing~\cite{OpenAI2023GPT4TR,anil2023palm,qiu2024towards, wu2024pmc} and computer vision~\cite{li2023blip,alayrac2022flamingo,gpt-4v-system-card, zhang2023pmc,wu2023can, zhou2024knowledge}, the pursuit of generalist medical artificial intelligence has grown increasingly promising and appealing~\cite{moor2023foundation,Wu2023TowardsGF,tu2024towards,zheng2023large, zhao2023one}, leading to the development of free-text generative foundation models capable of understanding and interpreting radiology imaging studies.
Yet, the complexity and specialized nature of clinical free-form texts, such as radiology reports and discharge summaries, present substantial challenges in evaluating model performance.
There is an urgent need for a robust and lightweight free-text evaluation metric to better monitor the development of medical generative foundation models and drive advancements in the field of generalist medical artificial intelligence.

\begin{figure}[!t]
    \centering
    \includegraphics[width=1\linewidth]{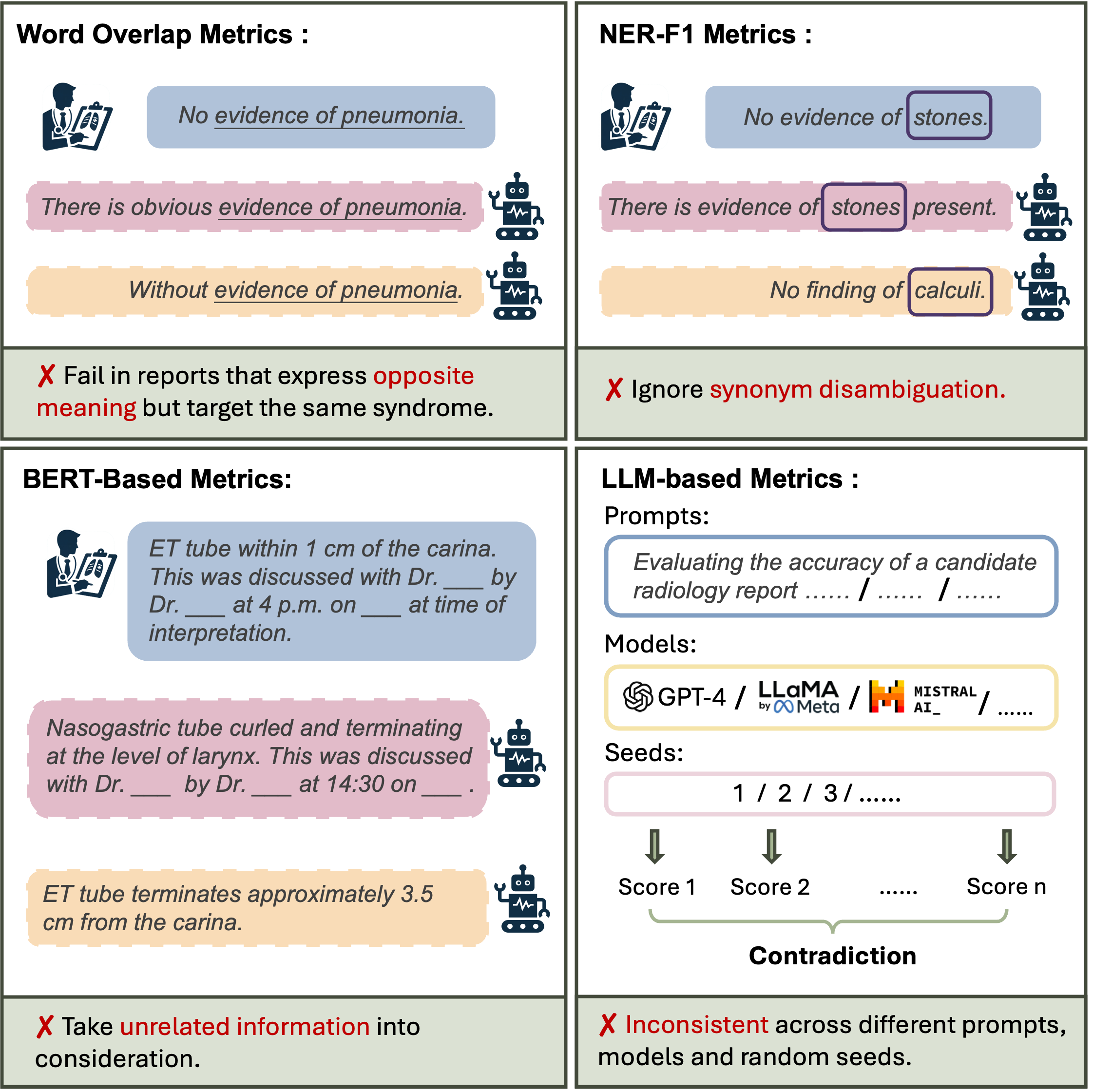}
    \caption{\textbf{Existing evaluation metrics.} We illustrate the limitations of current metrics. Blue boxes represent ground-truth reports; red and yellow boxes indicate correct and incorrect generated reports, respectively. 
    The examples show that these metrics fail to identify opposite meanings and synonyms in the reports and are often disturbed by unrelated information.\vspace{-15pt}}
    \label{fig:teaser}
    \vspace{-5pt}
\end{figure}

In the existing literature, four main types of metrics have been adopted to assess the similarity between free-form texts in medical scenarios, as shown in Figure~\ref{fig:teaser}. 
These include: 
(i) Word Overlap Metrics, such as BLEU~\cite{papineni2002bleu} and ROUGE~\cite{lin2004rouge}. While intuitive, these metrics fail to capture negation or synonyms in sentences, thus neglecting the semantic factuality; 
(ii) Embedding Similarity Metrics, like BERTScore~\cite{zhang2019bertscore},
provide better semantic awareness but fail to emphasize key medical terms, leading to overlooking errors in critical conclusions; 
(iii) Metrics based on Named Entity Recognition (NER), such as RadGraph F1~\cite{yu2023evaluating} and MEDCON~\cite{yim2023aci}. 
Although tailored for the medical domain, they struggle with synonym unification and are typically restricted to analyzing Chest X-ray reports;
(iv) Metrics relying on large language models (LLMs)~\cite{wei2024long,liu2023g}. 
While these metrics are better aligned with human preferences, they suffer from potential subjective biases and are
prohibitively costly for large-scale evaluation.

In this study, we aim to develop a metric that prioritizes key medical entities such as diagnostic outcomes and anatomical features, 
while exhibiting robustness against complex medical synonyms and sensitivity to negation expressions. We present two contributions: 
First, we introduce \textbf{RaTEScore}, a novel evaluation metric specifically designed for radiology reports. This metric emphasizes entity-level assessments across various imaging modalities and body regions. 
Specifically, it starts by identifying medical entities and their types ({\em e.g.}, anatomy, disease) from each sentence. To address the challenges associated with medical synonyms, we compute entity embeddings using a synonym disambiguation module and assess their cosine similarities. RaTEScore then calculates a final score based on weighted similarities that emphasize the clinical importance of the entity types involved.

Second, we have developed a comprehensive medical named-entity recognition (NER) dataset, \textbf{RaTE-NER}, which encompasses data from 9 modalities and 22 anatomical regions, derived from MIMIC-IV and Radiopaedia. 
In addition, we introduce \textbf{RaTE-Eval}, a novel benchmark designed to assess metrics across various clinical texts. This benchmark is structured around three sub-tasks: Sentence-level Human Rating, Paragraph-level Human Rating, and Comparison of Synthetic Reports, each targeting specific evaluation challenges. Both the RaTE-NER dataset and the RaTE-Eval benchmark have been made publicly available, aiming to foster the development of more effective evaluation metrics within the field of medical informatics.

Our extensive experiments demonstrate the superiority of \textbf{RaTEScore}. We initially tested our metric against others on the public dataset ReXVal~\cite{yu2023evaluating}, and it shows superior alignment with human preference. Considering ReXVal’s limitation to chest X-ray reports, further testing was conducted on the diverse sub-tasks of RaTE-Eval, where RaTEScore consistently outperformed other metrics. We also conduct ablation studies to validate the effectiveness of different individual components in our pipeline.

\section{Methods}

In this section, we start by properly formulating the problem, 
and introducing the pipeline of our metric~(Sec.~\ref{subsec:pipelie}).
Then, we detail each of the modules for our metric computation,
\emph{i.e.}, medical named entity recognition~(Sec.~\ref{subsec:ner}),
synonym disambiguation encoding~(Sec.~\ref{subsec:syn}),
and the final scoring procedure~(Sec.~\ref{subsec:score}).
Lastly, we present the implementation details at each stage.

\begin{figure*}[!t]
    \centering
    \includegraphics[width=1\linewidth]{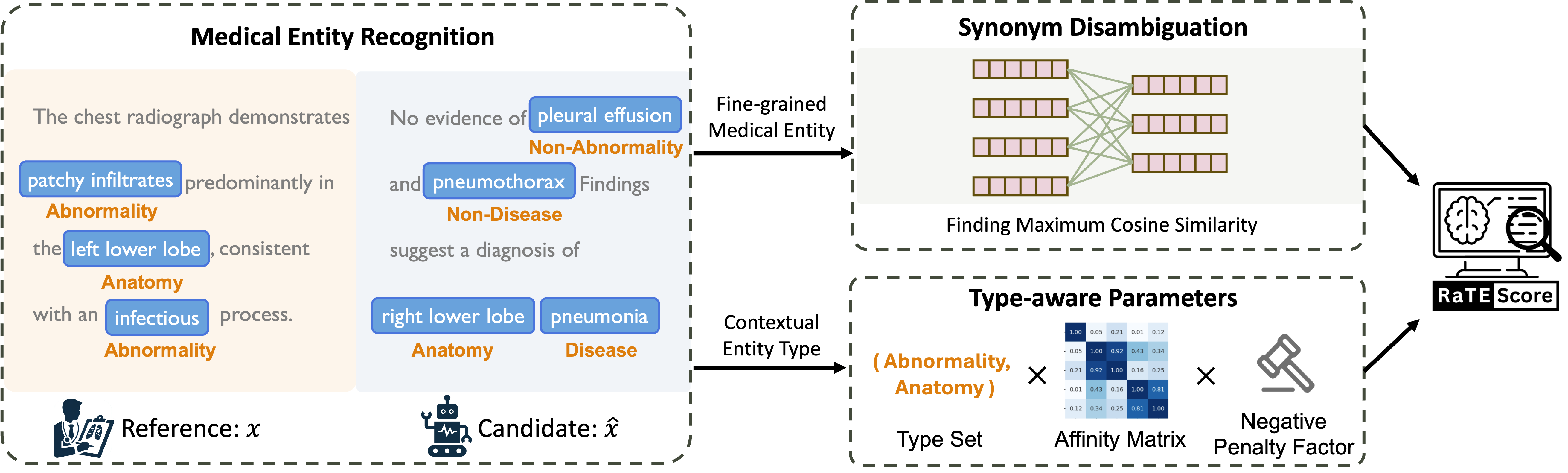}
    \caption{\textbf{Illustration of the Computation of RaTEScore.} Given a reference radiology report $x$, a candidate radiology report $\hat{x}$, we first extract the medical entity and the corresponding entity type. Then, we compute the entity embedding and find the maximum cosine similarity. The RaTEScore is computed by the weighted similarity scores that consider the pairwise entity types.\vspace{-10pt} }
    \label{fig:model}
\end{figure*}

\subsection{General Pipeline}
\label{subsec:pipelie}

Given two radiological reports, one is the ground truth for reference, denoting as $x$, and the other candidate for evaluation as $\hat{x}$. 
We aim to define a new similarity metric $S(x,\hat{x})$, that enables comparison of two radiological reports at the entity level, thus better reflecting their clinical consistency.

As shown in Figure~\ref{fig:model},  our pipeline contains three major components:
namely, a medical entity recognition module~($\Phi_\text{NER}(\cdot)$), 
a synonym disambiguation encoding module~($\Phi_\text{ENC}(\cdot)$), 
and a final scoring module~($\Phi_\text{SIM}(\cdot)$).
First, we extract the medicial entities from  each piece of radiological text, then encode each entity into embeddings that are aware of medical synonym, formulated as:
\begin{equation}
    \bold{F} = \Phi_\text{ENC}(\Phi_\text{NER}(x))
\end{equation}
where $\bold{F}$ contains a set of entity embeddings. 
Similarly, we can get $\bold{\hat{F}}$ for $\hat{x}$. 
Then, we can calculate the final similarity on the entity embeddings as:
\begin{equation}
    S(x,\hat{x}) = \Phi_\text{SIM}( \bold{F}, \bold{\hat{F}})
\end{equation}
In the following sections, we will detail each of the components. 

\subsection{Medical Named Entity Recognition}
\label{subsec:ner}

In the medical named entity recognition module, our goal is to decompose each radiological report by identifying a set of entities:
\begin{align*}
    \Phi_\text{NER} (x) &= \{ e_1, e_2, \dots, e_M \}\\
    &= \{(n_1, t_1), (n_2, t_2), \dots (n_M, t_M)\}
\end{align*}
Similarly, we can also get $\Phi_\text{NER} (\hat{x}) = \{ \hat{e}_1, \hat{e}_2, \dots, \allowbreak \hat{e}_N \}$, where $M, N$ denote the total number of entities extracted from each text respectively. Each entity $e_i$ is defined as a tuple $(n_i, t_i)$, where $n_i$ is the name of the entity and $t_i$ denotes its corresponding type. For instance, the tuple (`pneumonia', `Disease') represents the entity `pneumonia' categorized under the entity type `Disease'. 

Overall, we categorize the entity types into five distinct groups within radiological contexts: \{\textit{Anatomy, Abnormality, Disease, Non-Abnormality, Non-Disease}\}. Specifically, `Abnormality' refers to notable radiological features such as masses, effusion, and edema. Conversely, `Non-Abnormality' denotes cases where such abnormalities are negated in the context, as illustrated by the classification of `pleural effusion' in the statement `No evidence of pleural effusion'.
Compared to `Abnormality', `Disease' in radiology reports are more high-level, mainly about final main professional diagnosis conclusions, terms such as `pneumonia' or `lymphadenopathy'.

\vspace{-5pt}

\begin{table}[t]
\centering
\fontsize{7.5}{11.2}\selectfont
\begin{tabular}{rcccc}
\toprule
Data Source       & Train Set          & Dev Set        & Test Set        \\ \midrule
MIMIC-IV          & 10588        & 1323            &       1324           \\
Radiopaedia          & 30005     & 3600      & 3529    \\ \midrule
\rowcolor[HTML]{EFEFEF} 
\textbf{Total Reports}     & \textbf{40593}       & \textbf{4923 }           &     \textbf{  4853 }      \\ \bottomrule
\end{tabular}
\caption{\footnotesize \textbf{RaTE-NER Dataset Statistics (Report-level)}: 
The number of sentences from medical reports of each data source.}

\label{table:data_2}

\end{table}

\vspace{6pt}\noindent \textbf{RaTE-NER Dataset.} 
To support the development of our medical entity recognition module, 
we have constructed the \textbf{RaTE-NER} dataset, a large-scale, radiological named entity recognition (NER) dataset, including 13,235 manually annotated sentences from 1,816 reports within the MIMIC-IV database, that spans 9 imaging modalities and 23 anatomical regions, ensuring comprehensive coverage. 
Given that reports in MIMIC-IV are more likely to cover common diseases,
and may not well represent rarer conditions, we further enriched the dataset with 33,605 sentences from the 17,432 reports available on Radiopaedia~\cite{Radiopaedia}, by leveraging GPT-4 and other medical knowledge libraries to capture intricacies and nuances of less common diseases and abnormalities. More details can be found in the Appendix~\ref{sec:anno}. 
We manually labeled 3,529 sentences to create a test set.
As shown in Table~\ref{table:data} and Table~\ref{table:data_2}, the \textbf{RaTE-NER} dataset offers a level of granularity not seen in previous datasets, with comprehensive entity annotations within sentences, that enables to train models for medical entity recognition within our analytical pipeline.

\begin{table}[t]
\centering
\fontsize{6.5}{11.2}\selectfont
\begin{tabular}{rcccc}
\hline
                  & \multicolumn{3}{c}{\cellcolor[HTML]{EFEFEF}\textbf{MIMIC-IV}}    \\ \cline{2-4} 
                  & Train Set          & Dev Set        & Test Set        \\ \hline
Anatomy        &  9034 (4314)              &   1188 (828)              &        1140 (765)        \\
Abnormality    &   5579 (4047)                &  760 (657)                &      605  (513)         \\
Non-Abnormaliy &    4182  (1528)             &  479  (274)              &       514   (253)       \\
Disease        &   1675  (1220)              &  189 (169)               &       178  (164)        \\
Non-Disease    &      3482 (965)            &  424 (268)                &       457  (264)        \\ \hline
                  & \multicolumn{3}{c}{\cellcolor[HTML]{EFEFEF}\textbf{Radiopaedia}} \\ \cline{2-4} 
                  & Train Set          & Dev Set        & Test Set        \\ \hline
Anatomy        & 34110 (14051)       & 4145 (2629)            & 4471 (2889)          \\
Abnormality    & 33863 (23352)       & 4021 (3386)           & 4265 (3365)         \\
Non-Abnormaliy & 3878 (2280)         & 473 (325)           & 605 (420)            \\
Disease        & 9639 (7385)         & 1118 (1044)            & 741 (659)            \\
Non-Disease    & 2467 (1540)         & 268 (220)             & 183 (142)           \\ \midrule
\rowcolor[HTML]{EFEFEF} 
\textbf{Total Entities}    &      \textbf{   107909 (60682)      }    &     \textbf{   13065 (9800)      }    &      \textbf{  13159 (9434)    }     \\ 

\bottomrule
                  
\end{tabular}
\caption{\footnotesize \textbf{RaTE-NER Dataset Statistics (Entity-level)}: 
The numbers outside and inside the brackets denote the total number of entities for each type, and the number of non-duplicate entities, respectively. }
\label{table:data}
\end{table}

\vspace{-5pt}
\subsection{Synonym Disambiguation Encoding}
\label{subsec:syn}
To address the challenge from synonyms in the evaluation process, 
such as reconciling terms like ``lung'' and ``pulmonary'', 
we propose to map each entity name into embedding space, where synonyms are positioned closely together, utilizing a medical entity encoding module trained with extensive medical knowledge. 
This module, represented as: $f_i = \Phi_\text{ENC}(n_i)$, 
with $f_i$ denotes the vector embedding for the entity name. 
Consequently, we compile these into a set of entity embeddings: 
$\bold{F} = \{(f_1, t_1), (f_2, t_2), \dots\}$.
A similar set, \( \bold{\hat{F}} \), is constructed for the candidate text using the same methodology. For this encoding process, 
We adopt an off-shelf retrieval model, namely, BioLORD~\cite{remy2024biolord}, which is trained specifically on medical entity-definition pairs and has proven effective in measuring entity similarity.

\subsection{Scoring Procedure}
\label{subsec:score}
Upon obtaining the encoded entity set from each decomposed radiological report, 
we proceed to the final scoring procedure. We first define the similarity metric between a candidate entity and a reference report, that is established by selecting an entity from the referenced text based on the cosine similarity of their name embeddings:
\begin{equation}
i^* = \arg\max_{i \leq M} \cos(f_i, \hat{f}_j)
\end{equation}
where $\cos(f_i, \hat{f}_j)$ measures the cosine similarity between two entity name embeddings. The entity $e_{i^*}$, which best matches $\hat{e}_j$ from the reference text, is chosen for further comparison.
The overall similarity score, $S(x, \hat{x})$, is then computed as follows:
\vspace{-5pt}
\begin{equation}
    S(x, \hat{x}) = \frac{\sum_j{W(t_{i^*}, t_j}) \cdot \text{sim}(e_{i^*}, \hat{e}_j) }{\sum_j{W(t_{i^*}, t_j)}}
\end{equation}
Here, $W$ is a learnable $5 \times 5$ affinity matrix between the five entity types, where $W(t_i, t_j)$ represents an element of the matrix, 
and $\text{sim}(e_{i}, \hat{e}_j)$ is an entity-wise similarity function, defined as:
\vspace{-5pt}
\begin{equation}
    \text{sim}(e_{i}, \hat{e}_j) = \left \{
     \begin{aligned}
      {p} \cos(f_{i}, \hat{f}_j), \quad if \quad t_{i} \neq t_j \\
     \cos(f_{i}, \hat{f}_j), \quad if \quad t_{i}=t_j
     \end{aligned}
     \right. 
\end{equation}
where we generally follow the cosine similarity on the name embedding,
with a learnable penalty value $p$ to punish the type mismatch.
For example, when comparing entities with identical names but different types—such as (`pleural effusion', `Abnormality')  and (`pleural effusion', `Non-Abnormality')—the penalty term $p$ is applied to adjust the similarity score appropriately.

Additionally, the similarity between different entity types may be weighted differently in medical scenarios due to their clinical significance. For example, the similarity between two `Abnormality' entities is of much greater importance than the similarity between two `Non-abnormality' entities. This is because all body parts are assumed to be normal in radiology reports by default, and minor expression errors in normal findings will not critically impact the report's correctness. Therefore, we introduce $W$ to account for this clinical relevance.

Finally, due to the order of performing max indexing for selecting referenced entities and weighted sum pooling on all candidate entities, the final similarity metric $S(x,\hat{x})$ does not comply with the commutative law. $S(x,\hat{x})$ and $S(\hat{x},x)$ can be analogous to precision and recall respectively. Thus, our final \textbf{RaTEScore} is defined as the harmonic mean of $S(x,\hat{x})$ and $S(\hat{x},x)$, following classical F$_1$-score format:
\vspace{-5pt}
\[
\texttt{RaTEScore} = 
\begin{cases} 
0,\quad \; \  \text{if } S(x,\hat{x}) + S(\hat{x},x) = 0, \\
2 \times \frac{S(x,\hat{x}) \times S(\hat{x},x)}{S(x,\hat{x}) + S(\hat{x},x)}, \quad\  \text{otherwise}.
\end{cases}
\]

\vspace{-5pt}
\subsection{Implementation Details}
\label{subsec:imp}
In this section, we describe the implementation details for the three key modules. \textit{First}, for medical named entity recognition, 
we train a BERT-liked model on \textbf{RaTE-NER} dataset with two main-stream NER architectures, namely, Span-based and IOB-based models. 
For the Span-based method, we follow the setting of PURE~(the Princeton University Relation Extraction system) entity model~\cite{zhong2020frustratingly} and for the IOB-based method, 
we follow DeBERTa v3~\cite{he2022debertav3, he2021deberta}. 
We show more detailed implementation parameters for the two training schemes in Appendix~\ref{sec:NER Details}. Additionally, we also try to initialize the NER model with different pre-trained BERT. More comparison of the two training schemes and different BERT initializations will be present in the ablation study. \textit{Second}, For the synonym disambiguation encoding, we directly use the off-shelf BioLORD-2023-C model version. Ablation studies are also conducted in Section~\ref{sec:exp}. \textit{Third}, for the final scoring module, we learn the affinity matrix $W$ and negative penalty factor $p$ leveraging TPE (Tree-structured Parzen Estimator)~\cite{bergstra2011algorithms} with a small set of human rating data. 

\begin{table*}[]
\centering
\fontsize{8.6}{11.2}\selectfont
\begin{tabular}{cccccccc}
\toprule
\multicolumn{2}{c}{}                & \textbf{Number} & \textbf{Type }  & \textbf{Scoring Principle  }            & \textbf{Data Source  }              & \textbf{Modality}           & \textbf{Anatomy}             \\ \midrule
\multicolumn{2}{c}{\textbf{ReXVal Dataset}}  & 200         & Sent. + Para. & Error Count                   & MIMIC\_CXR                 & 1 (X-ray)              & 1 (Chest)          \\\midrule
\multirow{3}{*}{\textbf{Ours}} & \textbf{Sent. level} & 2215        & Sent.         & Error Count / Potential Errors & \multirow{3}{*}{MIMIC\_IV} & \multirow{3}{*}{9} & \multirow{3}{*}{22} \\
                  & \textbf{Para. level} &   1856        & Para.         & 5-Point Scoring System         &                            &                    &                     \\
                      & \textbf{Sim. Report} & 847         & Sent.         & Mistral 8*7B                   &                            &                    & \\ \bottomrule                  
\end{tabular}
\caption{Comparison of RaTE-Eval Benchmark and existed radiology report evaluation Benchmark.}
\label{tab:benchmark}
% \vspace{-8pt}
\end{table*}

\vspace{-2pt}
\section{RaTE-Eval Benchmark}

To effectively measure the alignment between automatic evaluation metrics and radiologists' assessments in medical text generation tasks, 
we have established a comprehensive benchmark, \textbf{RaTE-Eval}, 
that encompasses three tasks, each with its official test set for fair comparison, as detailed below. The comparison of RaTE-Eval Benchmark and existed radiology report evaluation Benchmark is listed in Table~\ref{tab:benchmark}.

\vspace{3pt}\noindent\textbf{Sentences-level Human Rating.}
Existing studies have predominantly utilized the ReXVal dataset~\cite{yu2023radiology}, which requires radiologist annotators to identify and count errors in various potential categories. The metric's quality is assessed by the Kendall correlation coefficient between the total number of errors and the result from the automatic metric. 
The possible error categories are as follows:

\vspace{-0.15cm}
\begin{quote}
    1.~False prediction of finding; \\[2pt]
    2.~Omission of finding; \\[2pt]
    3.~Incorrect location/position of finding; \\[2pt]
    4.~Incorrect severity of finding; \\[2pt]
    5.~Mention the comparison that is absent from the reference impression; \\[2pt]
    6.~Omission of comparison describing a change from a previous study.
\end{quote}
\vspace{-0.12cm}

Building on this framework, we introduce two improvements to enhance the robustness and applicability of our benchmark:
\textbf{(1) normalization of error counts}: recognizing that a simple count of errors may not fairly reflect the informational content in sentences, we have adapted the scoring to annotate \textbf{the number of potential errors}. Specifically, we computed the sum of both correct and incorrect findings in the reference sentence. This sum represents the total number of potential errors that could occur. This approach normalizes the counts, ensuring a more balanced assessment across varying report complexities.
\textbf{(2) diversification of medical texts}: unlike existing benchmarks that are limited to Chest x-rays from the MIMIC-CXR dataset~\cite{johnson2019mimic}, our dataset includes \textbf{2215} reports spanning \textbf{9} imaging modalities and \textbf{22} anatomies from the MIMIC-IV dataset~\cite{johnson2020mimic}, the involving imaging modalities and anatomies are listed in Appendix~\ref{sec:inv}. 

Specifically, our annotation process is as follows: 
First, we divide the MIMIC-IV dataset into 49 subsets based on modality and anatomy. To conduct sentence-level evaluation, we split the report paragraphs into individual sentences by periods and remove the duplicates. 
Next, we randomly sample 25 sentences from each subset to create a reference report list and sample another 1000 reports to form a candidate report list. Subsequently, we use several evaluation metrics—BLEU, ROUGE, BERTScore, CIDEr, and our proposed RaTEScore to identify the most similar one from the candidate report list for each report in the reference report list. 
We take the union of all the metric results as the report pairs for manual annotations. Finally, each case in the annotation reports was annotated by two experienced radiologists with over five years of clinical practice. For each candidate report and the corresponding reference report, they are required to count errors in six provided categories as well as the number of potential errors, where the error refers to the candidate report's error based on the reference report. 

The final human rating result is computed by dividing the total error count by the the number of potential errors. The final sentence-level benchmark is composed of 
2215 reference report sentences, candidate report sentences and their rating score.
For parameter search~(Sec.~\ref{subsec:imp}), we divided all reports into a training set and a test set at an 8:2 ratio, to identify the most effective parameters that align with human scoring rules. 

\vspace{3pt}
\noindent\textbf{Paragraph-level Human Rating.} 
Given that medical imaging interpretation commonly involves the evaluation of lengthy texts rather than isolated sentences, we also incorporate paragraph-level assessments into our analysis, from the MIMIC-IV reports. 
However, as it is challenging for humans to completely count all errors in long paragraphs accurately, we established a 5-point scoring system for our evaluations, following the RadPEER~\cite{goldberg2017acr} - an internationally recognized standard for radiologic peer review. They defines "Concur with Interpretation" as "correct diagnosis". In our annotation process, due to the difficulty of counting errors in paragraph-level reports (too detailed and some errors are not of great clincial significance), we instructed the radiologists to approximate the ratio of correct diagnosess based on their clinical judgement, following the ‘RadPEER’ standard, which is more aligned with the human rating system for report writing in clinical.

The scores range from 5, denoting a perfectly accurate report, to 0, that indicates the report without any correct observations. Detailed scoring criteria are provided in Appendix~\ref{sec:rate}, guiding radiologists on how to assign scores at different levels. 

Specifically, our annotation process is as follows: 
first, we divide the MIMIC-IV dataset into 49 subsets based on modality and anatomy. Next, we sample 20 reports from each subset to create a reference list and 500 reports to form a candidate list. 
The report selection process is the same as sentence-level human rating. 
For each candidate report and the corresponding reference report, the radiologists are required to give a 5-point score.

The final benchmark in paragraph-level is composed of 1856 reference reports, candidate reports and their rating score. Similarly, for parameter search~(Sec.~\ref{subsec:imp}), we also divide all reports into training set and a test set at an 8:2 ratio.

\vspace{3pt} \noindent\textbf{Rating on the Synthetic Reports.}
Here, we aim to evaluate the sensitivity of our metric on handling synonyms and negations using synthetic data. 
Specifically, we employed Mixtral 8x7B~\cite{Jiang2024MixtralOE}, 
a sophisticated open-source Large Language Model (LLM), 
to rewrite \textbf{847} reports from the MIMIC-IV dataset. 
The rewriting was guided by two tailored prompts:
\begin{tcolorbox}[colback=red!2!white,colframe=red!50!black]
\footnotesize
  \textit{You are a specialist in medical report writing, 
  please rewrite the sentence, you can potentially change the entities into synonyms, but please keep the meaning unchanged.}
\end{tcolorbox}

On the other hand, opposite %\weidi{what do you mean by anonymous.....}\weike{fixed} 
reports were generated with:
% \vspace{-3pt}
\begin{tcolorbox}[colback=red!2!white,colframe=red!50!black]
\footnotesize
  \textit{You are a specialist in medical report writing, 
  please rewrite the following medical report to express the opposite meaning.}
\end{tcolorbox}
\vspace{-3pt}
This process results in a test set comprising triads of reports: the original, a synonymous version, and an anonymous version, detailed further in Appendix~\ref{sec:sim}. Ideally, effective evaluation metrics should demonstrate higher scores for synonymous reports compared to anonymous reports, thereby more accurately reflecting the true semantic content of the reports.

\section{Experiments}
\label{sec:exp}
\begin{figure*}[!t]
    \centering
    \includegraphics[width=0.95\linewidth]{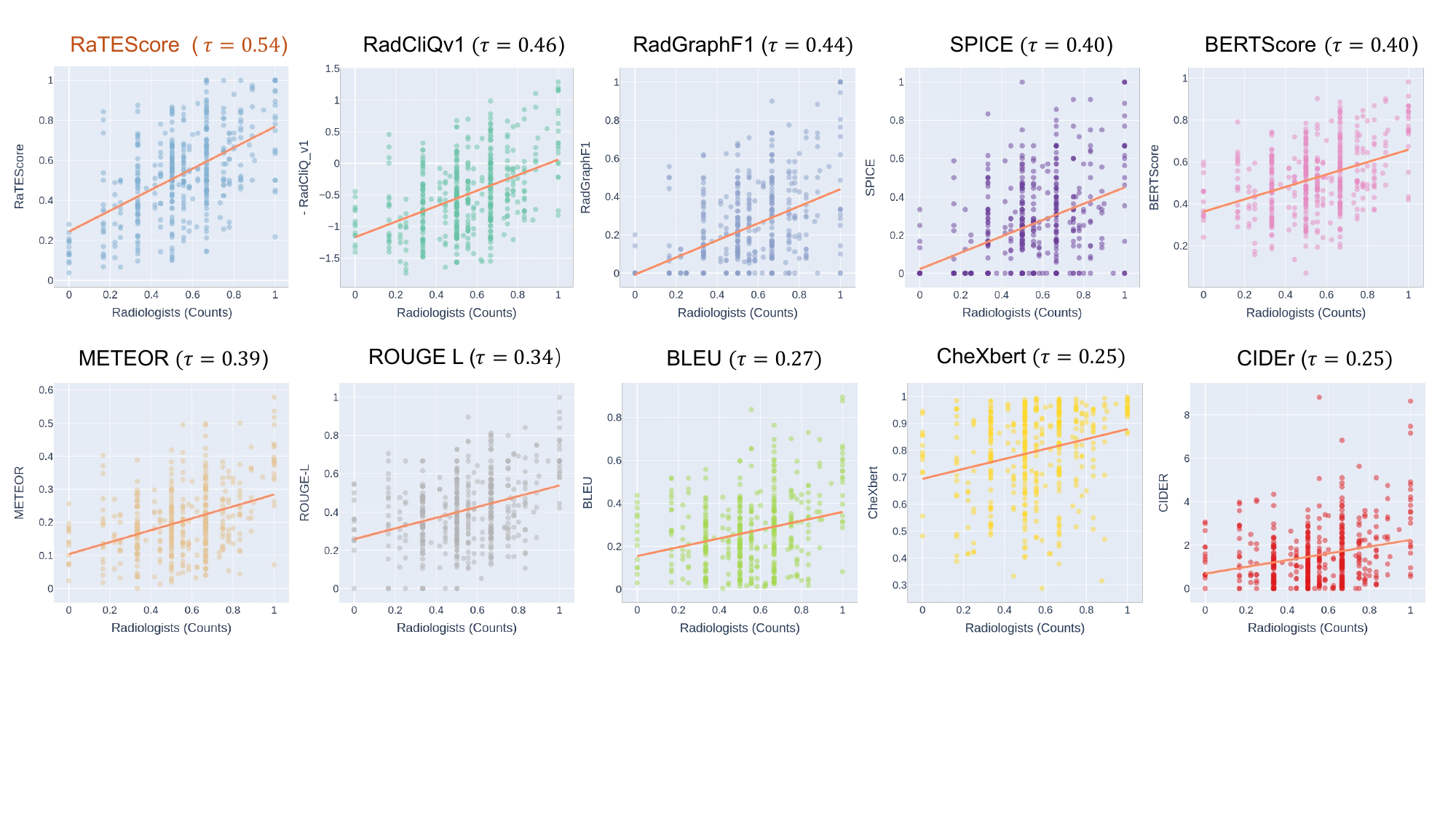}
    \vspace{5pt}
    \caption{\textbf{Results in RaTE-Eval Benchmark: Correlation Coefficients with Radiologists Results ( sentence-level ).} our metric exhibits the highest Pearson correlation coefficient with the radiologists' scoring. Note that the scores on the horizontal axis are experts counting various types of errors normalized by the potential error types that could occur in the given sentence, and subtracting this normalized score from 1 to achieve a positive correlation.}
    \label{fig:short}
    \vspace{-10pt}
\end{figure*}

In this section, we start by introducing the baseline evaluation metrics. 
Later, we compare the different metrics with our proposed RaTEScore on both ReXVal and RaTE-Eval benchmarks. Lastly, we present details for the ablation studies.

\vspace{-5pt}
\subsection{Baselines}
We use the following metrics as baseline comparisons:
BLEU~\cite{papineni2002bleu}, ROUGE~\cite{lin2004rouge}, 
METEOR~\cite{banerjee2005meteor}, CheXbert~\cite{smit2020chexbert,yu2023evaluating},
BERTScore~\cite{zhang2019bertscore}, SPICE~\cite{anderson2016spice} and RadGraph F1~\cite{yu2023evaluating}. Detailed explanations of these metrics can be found in the Appendix~\ref{sec:base}.

\subsection{Results in ReXVal dataset}

Our initial evaluation adopts the public ReXVal dataset, where we calculated the Kendall correlation coefficient to measure the agreement between our RaTEScore and the average number of errors identified by six radiologists. 
Our analysis was conducted under identical conditions to those of baseline methods. Given that the reports in ReXVal vary significantly in length, predominantly featuring longer documents, we applied a type weight matrix with parameters specifically fine-tuned on our long-report benchmark training set. As detailed in Table~\ref{table:result1}, RaTEScore demonstrated a Kendall correlation coefficient of $0.527$ with the error counts, surpassing all existing metrics.

\begin{table}[t]
\centering
\fontsize{8}{12}\selectfont
\setlength\tabcolsep{3.2pt}
\begin{tabular}{rccccc}
\toprule
& RadGraph F1 & BERTScore & CheXbert & BLEU & \textbf{Ours} \\
\cmidrule{2-6}
Kendall $\tau$ & 0.515*  & 0.511* & 0.499* & 0.462* & \textbf{0.527 }\\ \bottomrule
\end{tabular}
\caption{Results in ReXVal dataset: * denotes the result report in~\cite{yu2023evaluating}.}
\label{table:result1}
% \vspace{-5pt}
\end{table}

While further examining instances with notable deviations in Appendix~\ref{sec:fail}, a primary observation was that ReXVal's protocol tends to count six types of errors uniformly, without accounting for variations in report length. This approach leads to discrepancies where a single-sentence report with one error type and a twenty-sentence report with the same error count receive equivalent scores. To address this issue, our \textbf{RaTE-Eval} benchmark can be better suited to distinguish such variations, by normalising the total error counts with potential error counts.

\vspace{-5pt}
\subsection{Results in RaTE-Eval benchmark}

\noindent\textbf{On Sentence-level Rating.}
As illustrated in Fig.~\ref{fig:short}, our model achieved a Pearson correlation coefficient of $0.54$ on the RaTE-Eval short sentence benchmark, significantly outperforming the existing baselines. 
These results underscore the inadequacy of methods that predominantly rely on term overlap for evaluations within a medical context. While entity-based metrics like RadGraph F1 show notable improvements, they still do not reach the desired level of efficacy on an extensive benchmark encompassing multi-modal, multi-region reports. This shortfall largely attributes to the limited scope of training vocabulary in these methods.

\vspace{3pt} \noindent\textbf{On Paragraph-level Rating.}
From the results in Table~\ref{table:long}, it can be observed that \textbf{RaTEScore} shows a significantly higher correlation with radiology experts compared to other existing metrics, across various measures of correlation. Metrics that focus on identifying key entities, such as RadGraph F1, SPICE, and ours, consistently demonstrate stronger correlations than those reliant on mere word overlap, thereby supporting our primary assertion that critical statements in medical reports are paramount. Furthermore, metrics that accommodate synonyms, such as METEOR, outperform those that do not, such as BLEU and ROUGE. Significantly, \textbf{RaTEScore} benefits from a robust NER model trained on our comprehensive dataset, \textbf{RaTE-NER}, which spans multiple modalities and anatomical regions, not just Chest x-rays, resulting in markedly higher correlations.

\begin{table}[t]
\centering
\fontsize{8}{12}\selectfont
\setlength\tabcolsep{3.2pt}
\begin{tabular}{rcccc}
\toprule
\multicolumn{1}{l}{} & \multicolumn{3}{c}{\textbf{Paragraph-level Correlations}}         & \textbf{Simulations} \\ %\cline{2-5} 
                     & \textbf{Pearson $\tau$} & \textbf{Kendall $\tau$} & \textbf{Spearman $\tau$} & \textbf{Acc} \\ \midrule
RadGraph          & 0.624                   & 0.439                   & 0.582                    &      0.463       \\
BERTScore            & 0.599                   & 0.413                   & 0.555                    &   0.140          \\
CheXbert             & 0.496                   & 0.294                   & 0.403                    &   0.666       \\
BLEU                 & 0.409                   & 0.289                   & 0.404                    &   0.119       \\
ROUGE\_L             & 0.572                   & 0.396                   & 0.567                    &    0.117          \\
SPICE                & 0.623                   & 0.453                   & 0.605                    &    0.140          \\
METEOR               & 0.599                   & 0.422                   & 0.567                    &    0.168          \\\midrule
\textbf{Ours}        & \textbf{0.653}          & \textbf{0.462}          & \textbf{0.608}           &   \textbf{0.670}          \\ \bottomrule
\end{tabular}
\caption{Results in RaTE-Eval Benchmark: Correlation coefficients with radiologists and accuracy for whether the synonym sentence can achieve higher scores than
the antonymous one on Synthetic Reports.
}
\label{table:long}
% \vspace{-8pt}
\end{table}

\vspace{3pt} \noindent\textbf{Results on Synthetic Reports.}
To further showcase the effectiveness of our proposed \textbf{RaTEScore}, 
we examined its performance on the synthetic test set. This dataset, being synthesized, allows us to use accuracy (ACC) as a measure to evaluate performance. Specifically, we assess whether the synonymously simulated sentences received higher scores than their antonymous counterparts. The results, presented in Table~\ref{table:long}, demonstrate that our model excels in managing synonym and antonym challenges, affirming its robustness in nuanced language processing within a medical context.

\vspace{-4pt}
\subsection{Ablation Study}
In this ablation section, we investigate the pipeline from two aspects: 
namely, the design of NER model, the effect of different off-the-shelf synonym disambiguation encoding module.

\subsubsection{NER Module Discussion}
Here, we discuss the performance of our NER module in three parts: training schemes, initialization models, and data composition.

\vspace{3pt}
\noindent\textbf{Training Schemes.}
To select the most suitable NER model for training, we compare IOB-based and Span-based NER training schemes on the whole RaTE-NER test set. As shown in Table~\ref{table:ab1}, the IOB scheme overall extracts more comprehensive entities, but the recall is lower against the Span-based approach.

\vspace{3pt}
\noindent\textbf{Initialization Models.}
Additionally, as shown in Table~\ref{table:ab1}, we also try a sequential pre-trained BERT model for initialization, \emph{i.e.}, DeBERTa\_v3~\cite{he2022debertav3}, Medical-NER~\cite{MedicalNER}, BioMedBERT~\cite{Chakraborty2020BioMedBERTAP}, BlueBERT~\cite{peng2019transfer}, MedCPT-Q-Enc.~\cite{jin2023medcpt}, and BioLORD-2023-C~\cite{remy2024biolord}. 
Detailed description for each model can be found in Appendix~\ref{sec:bert}.  We apply various models in different training schemes based on their pre-training tasks. For example, Medical-NER is pre-trained with IOB-based NER tasks on other tasks thus we still finetune it in the same setting.
Comparing Medical-NER and DeBERTa\_v3, pretraining on other NER datasets does not improve much. Different types of BERT also perform fairly for the Span-based method. Based on the results, our final scores are all based on the IOB scheme with DeBERTa\_v3.

\begin{table}[t]
\vspace{-3pt}
\fontsize{8}{12}\selectfont
\setlength\tabcolsep{3.2pt}
\begin{tabular}{clcccc}
\toprule
                            & \textbf{Initialized BERT}  & \textbf{Pre} & \textbf{Recall} & \textbf{F1}& \textbf{Acc} \\ \midrule
\multirow{2}{*}{\textbf{IOB.}}  & DeBERTa\_v3        &\textbf{0.567}  & 0.575 & 0.571& 0.754\\
 & Medical-NER  &0.559  & 0.572 & 0.565& \textbf{0.759} \\ \midrule
\multirow{5}{*}{\textbf{Span.}} & BiomedBERT        & 0.556 &  0.676 & \textbf{0.610 }  & 0.730\\
                            & SapBERT            &   0.560 & 0.658 & 0.605  & 0.731\\
                            & BlueBERT    & 0.554 & 0.657 & 0.601 & 0.726 \\
                            & MedCPT-Q-Enc.    &0.470    &  \textbf{0.682} & 0.556 &0.678  \\
                            & BioLORD-2023-C    &   0.555 &  0.664   & 0.605       & 0.727  \\ \bottomrule
\end{tabular}
\caption{Ablation Study on NER Model Schemes. \vspace{-8pt}}
% \vspace{-8pt}
\label{table:ab1}
\end{table}
\vspace{5pt}

\noindent \textbf{Data Ablation.}
Our RaTE-NER data is composed of two distinct parts, and we conducted experiments to highlight the necessity of both. As shown in Table~\ref{table:ab2}, `R.' denotes data from Radiopaedia, 
while `M.' refers to the data from MIMIC-IV. 
By combining these two parts (denoted as `R.+M.'), 
we observe a significant improvement in the final NER performance, 
with an increase of 0.030 in F1 and 0.010 in ACC. 
This underscores the importance of incorporating each dataset component.
\begin{table}[h]
% \vspace{-8pt}
\fontsize{8}{12}\selectfont
\setlength\tabcolsep{8pt}
\begin{tabular}{ccccc}
\toprule
\textbf{Training Data}  & \textbf{Pre} & \textbf{Recall} & \textbf{F1} & \textbf{Acc} \\ \midrule
R.            &  0.525 & 0.558 &  0.541 & 0.727 \\
M.            & 0.515 & 0.550 &  0.531 & 0.744  \\ \midrule
\textbf{R. + M. }         &\textbf{0.567}  & \textbf{0.575} &   \textbf{0.571}& \textbf{0.754}    \\ \bottomrule
\end{tabular}
\caption{Ablation Study on NER Training Data. R. denotes data from Radiopaedia and M. denotes data from MIMIC-IV.}
% \vspace{-8pt}
\label{table:ab2}
\end{table}
\vspace{-8pt}

% \vspace{-10pt}
\subsubsection{Entity Encoding Module Discussion}

For the evaluation of Entity Encoding Module, we compare several off-the-shelf entity encoding models trained using different approaches on the sentence-level correlation task of RaTE-Eval. 
BioLORD-2023-C~\cite{remy2024biolord} is trained on medical entity concepts, MedCPT-Query-Encoder~\cite{jin2023medcpt} is trained on PubMed user click search logs, while RadBERT~\cite{chambon2023improved}, CXR-BERT~\cite{2204_09817}, and BioViL-T~\cite{bannur2023learning} are pre-trained on a large corpus of radiology texts. As shown in Table~\ref{table:ab3}, BioLORD, due to its original training goal covering medical entity normalization, which aligns with our needs in the Entity Encoding module, achieved the best performance. 
Based on these results, we selected BioLORD-2023-C as the base model for our Entity Encoding Module.

\begin{table}[h]
% \vspace{-8pt}
\fontsize{7}{9}\selectfont
\setlength\tabcolsep{3.2pt}
\begin{tabular}{cccccc}
\toprule
& BioLORD & MedCPT &RadBERT&	CXR-BERT&	BioViL-T \\ \midrule
Pearson $\tau$ & \textbf{0.540}  &  0.498 & 0.519 & 0.368 & 0.465    \\ \bottomrule   
\end{tabular}
\caption{Ablation Study on Pretrained Model of Entity Encoding Module.}
% \vspace{-8pt}
\label{table:ab3}
\end{table}

\vspace{-3pt}
\section{Related Work}
\vspace{-2pt}

\vspace{-2pt}
\subsection{General Text Evaluation Metric }
Automated scoring methods allow for a fair evaluation of the quality of generated text. 
BLEU~\cite{papineni2002bleu}, ROUGE~\cite{lin2004rouge} was originally designed for machine translation tasks, focusing on word-level accuracy. 
METEOR~\cite{banerjee2005meteor} adopts a similar design, taking into account synonym matching and word order. 
SPICE~\cite{anderson2016spice} uses the key objects, attributes, and their relationships to compute the metric. BERTScore~\cite{zhang2019bertscore}, a model-based method, assigns scores to individual words and averages these scores to evaluate the text's overall quality, facilitating a more detailed analysis of each word's contribution.
\vspace{-2pt}
\subsection{Radiological Text Evaluation Metric}

With the advancement of medical imaging analysis, researchers have recognized the importance of evaluating the quality of radiology text generation. 
Metrics such as CheXbert F1~\cite{smit2020chexbert} and RadGraph F1 ~\cite{yu2023evaluating} are based on medical entity extraction models. However, CheXbert can only annotate 14 chest abnormalities, and RadGraph F1~\cite{jain2021radgraph} is only trained on chest X-ray modality. 
MEDCON~\cite{yim2023aci} expands the extraction range by QuickUMLS package~\cite{soldaini2016quickumls}, which relies on a string match algorithm that is not flexible. RadCliQ~\cite{yu2023evaluating} performs ensembling with BLEU, BERTScore, CheXbert vector similarity, and RadGraph F1 for a comprehensive yet less interpretable evaluation.
These metrics calculate the overlap between reference and candidate sentences while overlooking the issue of synonymy. 
Recently, metrics using Large Language Models~(LLMs) such as GPT-4, such as G-Eval~\cite{liu2023g}, LLM-as-a-Judge~\cite{zheng2024judging}, and LLM-RadJudge~\cite{wang2024llm} have emerged, closely mimic human evaluation levels.
However, these methods are unexplainable and may have potential subjective bias. Besides, their high computational cost also limits them for statistic robust large-scale evaluation.

\vspace{-2pt}
\subsection{Medical Named-Entity Recognition}

The MedNER task targets extracting medical-related entities from given contexts. Great efforts have been made in this domain~\cite{jin2023medcpt,monajatipoor2024llms,keloth2024advancing,li2023far,chen2023few}. 
Inspired by the success of this work, we believe MedNER models are strong enough to simplify and structure complex clinical texts, thus reducing the difficulty of automatically comparing two clinical texts. The most related work to ours is RadGraph~\cite{jain2021radgraph} which trained an NER model for Chest X-ray reports while we are targeting more general clinical reports regardless of their type. 

\vspace{-3pt}
\section{Conclusion}
\vspace{-2pt}
In this work, we propose a new lightweight, explainable medical free-form text evaluation metric, \textbf{RaTEScore}, by comparing two medical reports on the entity level. In detail, first, we build up a new medical NER dataset, \textbf{RaTE-NER} targeting a wide range of radiological report types and train a NER model on it. Then, we adopt this model to simplify the complex radiological reports and compare them on the entity embedding level leveraging an extra synonyms disambiguation encoding model. Our final RaTEScore correlates strongly with clinicians' true preferences, significantly outperforming previous metrics both on the former existing benchmark and our new proposed \textbf{RaTE-Eval}, while maintaining computational efficiency and interpretability.
\section*{Limitations}

Although our proposed metric, RaTEScore,  has performed well across various datasets, there are still some limitations. First, in the synonym disambiguation module, we evaluated the performance of several existing models and directly ultilized them without fine-tuning specifically for the evaluation scenario, which could be enhanced in the future. 
Furthermore, while we expanded from single-modality radiological report evaluation to multimodal whole-body imaging, we still only considered the issues within the radiological report scenario and did not extend to other medical contexts beyond radiology, nor to the evaluation of other medical tasks, like medical QA, summarization task. These areas require more exploration.
\section*{Acknowledgements}

This work is supported by the National Key R\&D Program of China (No. 2022ZD0160702).
\bibliography{custom}

\appendix

\clearpage

\section{Appendix}
\label{sec:appendix}

\subsection{Scoring Example}
In this section, we will show an example of calculating RaTEScore. Given a radiology report pair: 
\begin{tcolorbox}
[colback=red!2!white,colframe=red!50!black, width=.49\textwidth]
%\footnotesize
  \textbf{Referenced $x$}: A Foley catheter is in situ.\\
  \textbf{Candidate $\hat{x}$}: A Foley catheter is not in place.
\end{tcolorbox}
\noindent For simplicity, we will only describe the calculation procedure for $S(x, \hat{x})$ in text, and the calculation procedure for $S(\hat{x},x)$ is similar.

We first conduct \textbf{Medical Named Entity Recognition} to decompose the natural text into entities. For the referenced report, the entities list is: \{(``Foley catheter'', Anatomy), (``in situ'', Non-Abnormality) \} and for the candidate report is \{(``Foley catheter'', Anatomy), (``not in place'', Abnormality) \}.
Subsequently, these extracted entities are processed through the \textbf{Synonym Disambiguation Encoding Module}, which encodes the ``Foley catheter'' and ``in situ'' into feature embedding. Finally, during the \textbf{Scoring Procedure}, we pick out the most similar entity in the referenced report for each entity in the candidate report, \emph{i.e.}, ``Foley catheter'' paired with ``Foley catheter'' in the reference, and ``not in place'' with ``in situ''. Then, we get two cosine similarity scores based on the text embedding, 1.0 for ``Foley catheter'' and 0.83 for ``not in place''. The similarity score between (``not in place'', Abnormality) and (``in situ'', Non-Abnormality) will be further multiplied with a penalty factor $p$ as 0.37 while the other similarity is maintained since they have the same entity type. At Last, we calculate the weighted combination of the two. The weights are derived from a learnable attribution matrix $W$ corresponding to these type combinations, as 0.91 and 0.94 respectively. The calculation formulation is as follows:
\[
    \begin{aligned}
        S(x, \hat{x}) &= \frac{0.91\times1 + 0.94\times0.83\times0.36}{0.91+0.94} \\
          &= 0.644.
    \end{aligned}
\]
Similarly, we can get the other similarity:
\[
    \begin{aligned}
        S(\hat{x}, x) &= \frac{0.91\times1 + 0.83\times0.83\times0.36}{0.91 + 0.83} \\
          &= 0.666.
    \end{aligned}
\]
Notably, the only difference between the two similarity scores in this case lies in the weight between (``in situ'', Non-Abnormality) and (``not in place'', Abnormality). Due to the comparison directions, in $S(x, \hat{x})$, $W(\text{Non-Abnormality},\text{Abnormality})$ as 0.94 
is adopted and in the other hand, $W(\text{Abnormality},\text{Non-Abnormality})$ as 0.83 is adopted. The final score is computed as follows:
\[
\texttt{RaTEScore} = 2 \times \frac{{S(x,\hat{x})} \times {S(\hat{x},x)}}{{S(x,\hat{x})} + {S(\hat{x},x)}} = 0.676.
\]

\subsection{Automatic Annotation Approach}
\label{sec:anno}

\begin{figure*}[t]
    \centering
    \includegraphics[width=0.8\linewidth]{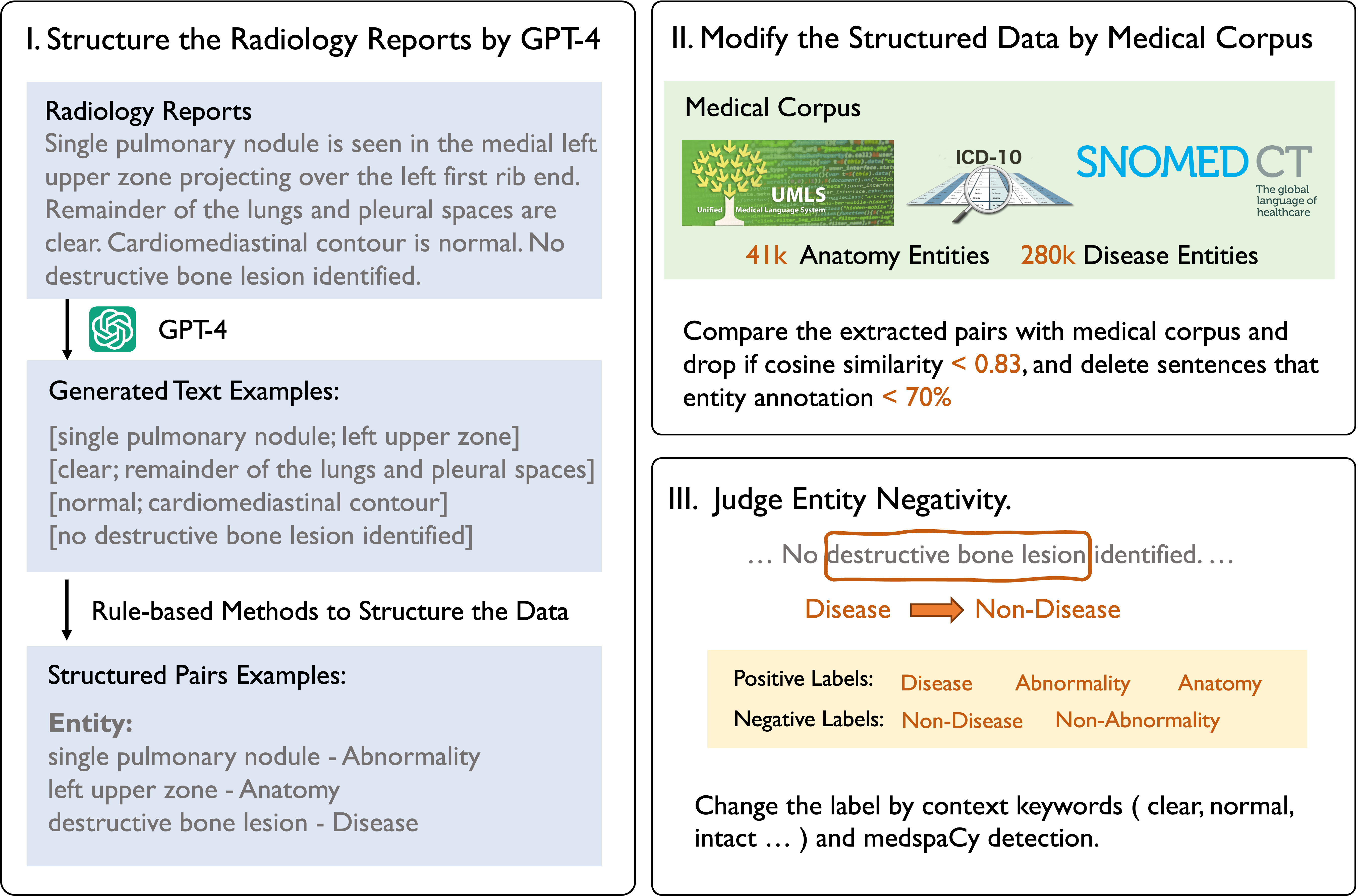}
    \caption{Data Curation Procedure.}
    \label{fig:data}
    % \vspace{-15pt}
\end{figure*}

Here, we introduce our automatic approach to construct a part of our \textbf{RaTE-NER} dataset, sourced from 19,263 original reports obtained from Radiopaedia~\cite{Radiopaedia} and covering 9 modalities and 11 anatomies. As shown in Figure~\ref{fig:data}, leveraging the latest LLM GPT-4 combined with other comprehensive medical knowledge bases, we develop a new automated medical NER and relation extraction dataset construction pipeline.

Specifically, we manually annotate several reports at the required granularity and adopt few-shot prompts with GPT-4 to initially establish an NER dataset. 

\begin{tcolorbox}
[colback=red!2!white,colframe=red!50!black, width=.49\textwidth]
  \textbf{GPT-4 prompt}: \\
  \textit{You are an AI assistant specializing in radiology reports reading. You are provided with a medical caption. Extract the entities and decide their type from organ, abnormal description or disease. Collect the organ and description together if the description modifies the organ. Leave disease alone. Make sure that the description is about the abnormality but not position. \\
  The output should follow this format: [organs; abnormal description] or [disease]. All words in [] should belong to the original sentence.}
\end{tcolorbox}

\begin{tcolorbox}
[colback=red!2!white,colframe=red!50!black, width=.49\textwidth]
%\footnotesize
  \textbf{Few-shot examples}: \\
  'context': \textit{"The sentence is: Hetergeneous and nodular enhancement of the liver with pre-contrast HU of -4 (!) indicating hepatic steatosis."}, \\
  'response': \textit{" [ liver; Hetergeneous and nodular enhancement ] [ liver; pre-contrast HU of -4 ] [ hepatic steatosis ] "}
\end{tcolorbox}

Following this, we build a robust medical entity library, integrating UMLS~\cite{bodenreider2004unified}, Snomed CT~\cite{donnelly2006snomed}, ICD-10~\cite{ICD10}, and other knowledge bases, then, compare all extracted entities using the MedCPT~\cite{jin2023medcpt} model for similarity. During the comparison process, entities with cosine similarity lower than 0.83 were filtered out. Most entities below this threshold did not meet our requirements. Subsequently, we removed sentences with an entity annotation density lower than 0.7 at the sentence level. Finally, we use medspaCy~\cite{medspacy} and also key negative words detection in reports, such as ``no'', ``without'', ``unremarkable'', ``intact'',
to determine the positive or negative polarity of each word in the sentence. 

\subsection{Involving Anatomies and Modalities in MIMIC-IV Data}
\label{sec:inv}
In this section, we detail the imaging modalities and anatomies involved in MIMIC-IV dataset.

\vspace{3pt}
\noindent \textbf{Anatomy List:} NECK, TEETH, BRAIN, HEAD, CHEST, PELVIS, ABDOMEN, CARDIAC, HEAD-NECK, SOFT TISSUE, UP-EXT, OB, EXT, HIP, BREAST, SPINE, MAMMO, BRAIN-FACE-NECK, LOW-EXT, BONE, VASCULAR, BLADDER.

\vspace{3pt}
\noindent \textbf{Modality List:} CT, CTA, Fluoroscopy, Mammography, MRA, MRI, MRV, Ultrasound, X-Ray.

\subsection{Guidelines for Radiologists}
\label{sec:rate}
Referencing RadPEER~\cite{goldberg2017acr}, we set up a five-point scoring criteria, as shown in Table~\ref{table:guide}. During the annotation process, each report is compensated with \$1 per report, with five reference reports separately.

\begin{table*}[t]
\fontsize{9.5}{13}\selectfont
\setlength\tabcolsep{3.2pt}
\begin{tabular}{clp{11cm}}
\toprule
\textbf{Score} & \textbf{Meaning  }            & \textbf{Explaination}                                                                                                                            \\ \midrule
5     & Correct              & Most of the diagnosis results are correct. Most descriptions are the same. Some wrong description unlikely to be clinically significant.            \\
4     & Almost Correct       & 75\% of the diagnosis results are correct. Most descriptions are the same. Some wrong description likely to be clinically significant. \\
3     & Partly Correct       & 50\% of the diagnosis results are correct.                                                                                                          \\
2     & Partly Incorrect     & 25\% of the diagnosis results are correct                                                                                                           \\
1     & Major Errors Present & Incorrect diagnosis. Maybe some negative descriptions are the same.                                                                                 \\
0     & Total Different      & No overlap for the descripted information.                                                                                                          \\ \bottomrule
\end{tabular}
\caption{5-point scoring system For Radiologists to Rate in Paragraph-level Human Rating of RaTE-Eval Benchmark}
\label{table:guide}
\end{table*}

\subsection{Example for Simulation Reports}
\label{sec:sim}
In this section, we give an example for the simulation report generation:
\begin{tcolorbox}[colback=red!2!white,colframe=red!50!black, width=.5\textwidth]
%\footnotesize
  \textbf{GT: }The appendix is well visualized and air-filled.\\
  \textbf{REWRITE: }The appendix is seen and contains gas.\\
  \textbf{OPPOSITE: }The appendix is poorly visualized and not air-filled.
\end{tcolorbox}

\subsection{Baselines}
\label{sec:base}
Herein, we will introduce the considered baselines:

\begin{itemize}
\setlength{\itemsep}{0pt}     
\setlength{\parsep}{0pt}     
\setlength{\topsep}{0pt}     
    \item BLEU~\cite{papineni2002bleu} measures the precision of generated text by comparing n-gram overlap between the generated report and  reference reports.
    \item ROUGE~\cite{lin2004rouge} focuses on the recall of generated text by measuring the overlap of n-grams, similar to BLEU. 
    \item METEOR~\cite{banerjee2005meteor} combines precision, recall, and a penalty for fragmented alignments, while also considering words order and synonyms through WordNet~\cite{fellbaum2010wordnet}.
    \item CheXbert~\cite{smit2020chexbert,yu2023evaluating} computes the cosine similarity between CheXbert model embedding of the reference report and candidate report.
    \item BERTScore~\cite{zhang2019bertscore} utilizes a pre-trained BERT model to calculate the similarity of word embeddings between candidate and reference texts.
    \item SPICE~\cite{anderson2016spice} extracts key objects, attributes, and their relationships from descriptions to build a scene graph, and compares the two texts on the scene graph level.
    \item RadGraph F1~\cite{yu2023evaluating} extracts the radiology entities and relations for Chest X-ray modality and computes the F1 score on the entity level.
\end{itemize}

\subsection{Failure Cases in ReXVal Dataset}
\label{sec:fail}

In this section, in order to better demonstrate the drawbacks of ReXVal dataset, we will give a failure case where two reports with different entity-wise errors while achieve the same scores.

\noindent \textbf{Report Pair 1:}
\begin{tcolorbox}[colback=red!2!white,colframe=red!50!black]
%\footnotesize
  \textbf{GT: } ET tube within 1 cm of the carina.    This was discussed with Dr. \_\_\_ at 4 p.m. on \_\_\_ by Dr. \_\_\_ at time of interpretation.
  
  \textbf{Pred: }ET tube terminates approximately 3 . 5 cm from the carina.

  \textbf{Total Errors:} 1.33
\end{tcolorbox}

\noindent \textbf{Report Pair 2:}
\begin{tcolorbox}[colback=red!2!white,colframe=red!50!black]
%\footnotesize
  \textbf{GT: }In comparison with the study of xxx, there is again enlargement of the cardiac silhouette with elevation of pulmonary venous pressure.  Opacification at the right base again is consistent with collapse of the right middle and lower lobes  RECOMMENDATION(S):  The tip of the right IJ catheter is in the mid to lower SVC.
    
  \textbf{Pred:} In comparison with the study xxx, there is little change in the appearance of the monitoring and support devices.  Continued substantial enlargement of the cardiac silhouette with relatively mild elevation of pulmonary venous pressure.  Opacification at the right base silhouettes the hemidiaphragm and is consistent with collapse of the right middle and lower lobes.

  \textbf{Total Errors:} 1.33
\end{tcolorbox}

As shown in the examples, case 1 with only two entity errors scores 1.3, and the report that describes more than ten different entity errors also scores 1.3. Moreover, reports length less than 10 words commonly has zero errors in ReXVal, whereas reports longer than 25 words had an average error count greater than 3, simply because the texts are longer and may contain more potential errors. Therefore, ignoring normalization and directly using absolute error counting numbers as the score like ReXVal may present severe bias that longer sentences scoring lower and shorter sentences scoring higher.

\subsection{Pretrained BERT Model Introduction}
\label{sec:bert}
In this section, we will introduce our considered pre-trained BERT models in detail:
\begin{itemize}
\setlength{\itemsep}{0pt}      % 去除项目间的间距
\setlength{\parsep}{0pt}       % 去除段落间的间距
\setlength{\topsep}{0pt}       % 去除列表顶部的间距
    \item DeBERTa\_v3~\cite{he2022debertav3} is an advanced version of the DeBERTa~\cite{he2021deberta} model, which improves upon the BERT and RoBERTa models by incorporating disentangled attention mechanisms, enhancing performance on a wide range of natural language processing tasks.
    \item Medical-NER~\cite{MedicalNER} is a fine-tuned version of DeBERTa to recognize 41 medical entities. The specific training data is not publicly available.
    \item BioMedBERT~\cite{Chakraborty2020BioMedBERTAP} previously named "PubMedBERT", pretrained from scratch using abstracts and full-text articles from PubMed~\cite{canese2013pubmed}.
    \item BlueBERT~\cite{peng2019transfer} is a BERT model pre-trained on PubMed abstracts and clinical notes (MIMIC-III)~\cite{johnson2016mimic}.
    \item MedCPT-Q-Enc.~\cite{jin2023medcpt} is pre-trained by 255M query-article pairs from PubMed search logs, and achieve SOTA performance on several zero-shot biomedical IR datasets.
    \item BioLORD-2023-C~\cite{remy2024biolord} is based on a sentence-transformers model and further finetuned on the entity-concept pairs.
\end{itemize}

\subsection{NER Module Implementation Details}
\label{sec:NER Details}
In the Medical Named Entity Recognition Module training scheme, 
we train the model on one NVIDIA GeForce GTX 3090 GPU with a batch size of 96 for 10 epochs while adopt different learning rates for different training schemes.  
For the Span-based method, we follow the setting of PURE entity model~\cite{zhong2020frustratingly}, which uses a pre-trained BERT model to obtain contextualized representations and then fed into a feedforward network to predict the probability distribution of the entity. It combines a BERT~\cite{devlin2018bert} model and a 3-layer MLP with head hidden dimension of 3096 for span classification. The span max length is 8. In the training stage, we set the learning rate as 6e-6. For the IOB-based method, each token is labeled as 'B-' (beginning of an entity), 'I-' (inside an entity), or 'O' (outside of any entity). We directly fine-tune the pre-trained BERT to perform a token classification task. Specifically, we add a linear layer to the output embedding of a BERT-liked model, which is fine-tuned utilizing a corpus of annotated entity data to predict the entity label for each token. We use a learning rate of 1e-5 for the IOB-based training scheme.

\end{document}